\documentclass[pdflatex,sn-mathphys,iicol]{sn-jnl}

\jyear{2025}

\theoremstyle{thmstyleone}

\theoremstyle{thmstyletwo}

\theoremstyle{thmstylethree}

\usepackage[T1]{fontenc}
\usepackage{lmodern}
\usepackage{fix-cm}
\usepackage{graphicx}
\usepackage{subcaption}
\usepackage{tabularx}
\usepackage{booktabs}
\usepackage{stfloats}
\usepackage{placeins}
\usepackage{amsmath}
\usepackage[table]{xcolor}

\usepackage{fvextra}

\DefineVerbatimEnvironment{PromptVerbatim}{Verbatim}{%
  fontsize=\normalsize,
  breaklines=true,
  breakanywhere=true,
  breaksymbolleft={},
  breaksymbolright={},
  frame=single,
  framerule=0.3pt,
  framesep=2pt,
  rulecolor=\color{black!55},
  numbers=none
}

\newcommand{\greyscalefig}[2][]{\includegraphics[#1]{#2}}
\newcommand{\compacttablefig}[2][1.10]{\makebox[\linewidth][c]{\includegraphics[width=#1\linewidth]{#2}}}

\makeatletter
\def\ps@plain{%
  \let\@oddhead\@empty
  \let\@evenhead\@empty
  \def\@oddfoot{\hfill\thepage\hfill}%
  \def\@evenfoot{\hfill\thepage\hfill}%
}
\def\ps@headings{%
  \let\@oddhead\@empty
  \let\@evenhead\@empty
  \def\@oddfoot{\hfill\thepage\hfill}%
  \def\@evenfoot{\hfill\thepage\hfill}%
}
\makeatother

\begin{document}

\title[Evaluation of Small Language Models for Arabic NLP]{Evaluation of Small Language Models for Arabic Language Processing}

\author{\fnm{Jumana} \sur{Alsubhi}}
\email{Jumana.Alsubhi@naseej.com}

\author{\fnm{Ahmed} \sur{Alhusayni}}
\email{Ahmad.Alhusainy@naseej.com}

\author{\fnm{Abdulrahman} \sur{Gharawi}}
\email{A.Gharawi@naseej.com}

\author{\fnm{Israa} \sur{Hamdine}}
\email{Israa.Hamdin@naseej.com}

\author{\fnm{Alshaymaa} \sur{Allahim}}
\email{A.Allahim@naseej.com}

\author{\fnm{Lamees} \sur{Alhumaid}}
\email{L.Alhumaid@naseej.com}

\author{\fnm{Ahmad} \sur{Shabana}}
\email{ashabana@naseej.com}

\author{\fnm{Rafik} \sur{Madani}}
\email{Madani.Rafik@naseej.com}

\affil{\orgname{Naseej Innovation Lab, Naseej for Technology, Riyadh}, \orgaddress{\country{Saudi Arabia}}}

\abstract{This paper evaluates the performance of twelve Small Language Models (SLMs) on Arabic natural language processing tasks. The study introduces a benchmark of 240 Arabic test items distributed across eight domains and ten language skills, covering both comprehension-oriented and generation-oriented tasks. All models were evaluated under a controlled zero-shot setting using a standardized Arabic-only prompt template. Model responses were assessed through a multi-model LLM-as-a-judge framework involving GPT-4.1 Mini, Claude Haiku 4.5, and DeepSeek-Chat, with scores aggregated across judges and analyzed by task, skill, and model family. The results show that Gemma 3 (12B) achieved the highest overall score (4.548/5), followed by Aya and C4AI Command Arabic. The observed results suggest that model size alone does not explain Arabic SLM performance. Models with stronger Arabic alignment and more reliable instruction-following behavior tended to perform better across tasks. Common failure patterns among lower-performing models include prompt leakage, hallucination, language drift, incomplete generation, and weak task adherence. Overall, the benchmark provides a structured reference for evaluating compact Arabic language models and supports future work on efficient, reliable, and culturally appropriate Arabic AI systems.}


\maketitle
\thispagestyle{plain}
\pagestyle{plain}

\section{Introduction}

Large Language Models (LLMs) have substantially advanced natural language processing (NLP), particularly in tasks involving language understanding, text generation, summarization, information extraction, reasoning, and dialogue systems \cite{ref6}. NLP is a field of artificial intelligence concerned with enabling computers to understand, interpret, generate, and interact with human language in both written and spoken forms. 
However, their practical deployment is often constrained by high memory requirements, computational cost, inference latency, and dependence on specialized hardware. These constraints are especially relevant in resource-limited environments, enterprise applications, edge deployments, and settings where predictable latency and operational cost are critical.

Small Language Models (SLMs) have therefore received increasing attention as efficient alternatives to large-scale systems. Small Language Models (SLMs) are language models designed with substantially fewer parameters than conventional large language models\cite{ref8}. While no universally accepted parameter threshold exists, SLMs are generally characterized by their ability to perform a wide range of language understanding and generation tasks while requiring significantly lower computational resources, memory consumption, and inference cost. Although they contain fewer parameters, recent SLMs can provide competitive performance when supported by appropriate pretraining, instruction tuning, alignment, and domain adaptation \cite{ref1}. Their reduced computational footprint makes them attractive for embedded systems, educational applications, institutional knowledge services, and other scenarios in which full-scale LLMs may be impractical.

The need for systematic SLM evaluation is particularly important for Arabic. Arabic is morphologically rich, highly inflectional, and characterized by substantial dialectal variation and orthographic ambiguity \cite{ref4, ref22}. These linguistic properties complicate both model training and evaluation, especially when models must follow instructions, preserve Arabic-only output, reason over context, and generate fluent responses across multiple domains. In addition, the availability of large, high-quality, and diverse Arabic evaluation resources remains more limited than for English and other high-resource languages.

The demand for reliable Arabic NLP systems is increasing across education, digital transformation, knowledge management, cultural preservation, government services, and enterprise automation \cite{ref5, ref6, ref7}. In this context, compact Arabic-capable models are important not only for research, but also for deployable Arabic AI systems, defined here as systems designed to process, understand, generate, or reason over Arabic-language content, while operating efficiently and respecting linguistic and cultural requirements.

This study evaluates twelve modern SLMs on a benchmark of 240 Arabic tasks distributed across eight domains and ten core language skills. The benchmark uses a standardized prompt design and a multi-model LLM-as-a-judge evaluation pipeline to compare model performance across comprehension and generation tasks. The study contributes: (i) a controlled benchmark design for Arabic SLM evaluation; (ii) a skill-level analysis of model strengths and weaknesses; (iii) an examination of evaluator disagreement; and (iv) a qualitative analysis of recurrent failure patterns affecting Arabic model reliability.

\section{Related Work}

Research on language-model evaluation has expanded rapidly with the emergence of instruction-tuned and generative LLMs. Early evaluation efforts primarily focused on task-specific performance, while more recent studies increasingly assess broader capabilities such as instruction following, reasoning, multilingual robustness, and open-ended generation quality. This shift is particularly relevant for SLMs, whose value depends on whether they can provide reliable performance under realistic deployment constraints.

\subsection{Small Language Models and Efficient NLP}

SLMs are designed to reduce the computational and operational cost of language-model deployment while retaining useful task performance. Prior work has shown that compact models can benefit significantly from supervised fine-tuning, instruction tuning, reinforcement learning from human feedback, and other alignment strategies \cite{ref8}. Open and multilingual model families such as Gemma, Qwen, Aya, Phi, and related compact architectures demonstrate that smaller models can perform competitively when trained on diverse data and optimized for instruction-following behavior \cite{ref8, ref12}. Nevertheless, performance is not determined by parameter count alone; training data quality, post-training procedures, and language coverage strongly influence model reliability.

\subsection{Arabic NLP and Arabic-Oriented Language Models}

Arabic NLP presents specific challenges because of the language's morphology, dialectal diversity, script variation, and context-dependent meaning \cite{ref16}. Earlier Arabic NLP research often emphasized task-specific systems for sentiment analysis, machine translation, named entity recognition, question answering, and text classification. Transformer-based Arabic models such as AraBERT marked an important step in adapting pretrained architectures to Arabic corpora and linguistic patterns \cite{ref17}. More recent multilingual and Arabic-oriented generative models have expanded the scope of Arabic NLP toward open-ended dialogue, instruction following, and text generation.

\subsection{Arabic Benchmarks}

Arabic benchmarking has also developed substantially. Benchmarks such as ArabicMMLU \cite{ref11}, ORCA \cite{ref13}, Dolphin \cite{ref18}, and DialectalArabicMMLU \cite{ref4} provide valuable resources for evaluating Arabic language understanding and generation across domains. These resources have improved the visibility of Arabic model performance and highlighted weaknesses in reasoning, dialectal understanding, and generation quality. However, many existing evaluations focus either on large proprietary models, broad multilingual systems, or isolated downstream tasks. As a result, there remains a need for unified evaluations of compact SLMs across diverse Arabic language skills under controlled prompting conditions.

\subsection{LLM-as-a-Judge Evaluation}

LLM-as-a-judge methods have become common for evaluating open-ended model outputs when automatic metrics are insufficient \cite{ref9,ref14}. These methods use strong language models to assess generated responses according to predefined rubrics such as correctness, clarity, completeness, and task relevance. Prior studies suggest that using multiple evaluators can reduce dependence on a single judge and provide more stable aggregate judgments \cite{ref14}. At the same time, judge models may differ in calibration and may reward different aspects of response quality, making disagreement analysis an important part of responsible evaluation \cite{ref10}.

\subsection{Research Gap}

Despite progress in Arabic language modeling and benchmarking, limited work has systematically evaluated Arabic-capable SLMs within a unified experimental framework that combines controlled prompts, multi-skill coverage, Arabic-only generation requirements, multi-model judging, and disagreement tracking. This study addresses that gap by evaluating modern SLMs across a structured Arabic benchmark and by analyzing both quantitative performance and qualitative failure patterns.

\section{Benchmark Construction}

The benchmark contains 240 Arabic test items designed to evaluate model capabilities across a broad set of domains and language skills. The dataset was constructed to provide balanced coverage across domains, with skill-level coverage designed to capture both comprehension- and generation-oriented behavior \cite{ref3,ref11,ref13}.

\subsection{Domain Coverage}

The benchmark includes eight domains, with thirty items in each domain. The selected domains were intended to reflect diverse contexts in which Arabic is used, including social topics, Arabic language analysis, history, technology, argumentative reasoning, religious content, mathematics, and creative writing \cite{ref1,ref16}. Table~\ref{tab:domains} summarizes the domain distribution.

\begin{table}[ht]
\centering
\begin{tabular}{lc}
\toprule
\textbf{Domain} & \textbf{Number of Questions} \\
\midrule
Social & 30 \\
Arabic Language & 30 \\
History & 30 \\
Technology & 30 \\
Argumentative & 30 \\
Religious & 30 \\
Mathematics & 30 \\
Creative Writing & 30 \\
\bottomrule
\end{tabular}
\caption{Distribution of benchmark questions across domains.}
\label{tab:domains}
\end{table}

\subsection{Skill Coverage}

\begin{figure}[ht]
\centering
\greyscalefig[width=\linewidth]{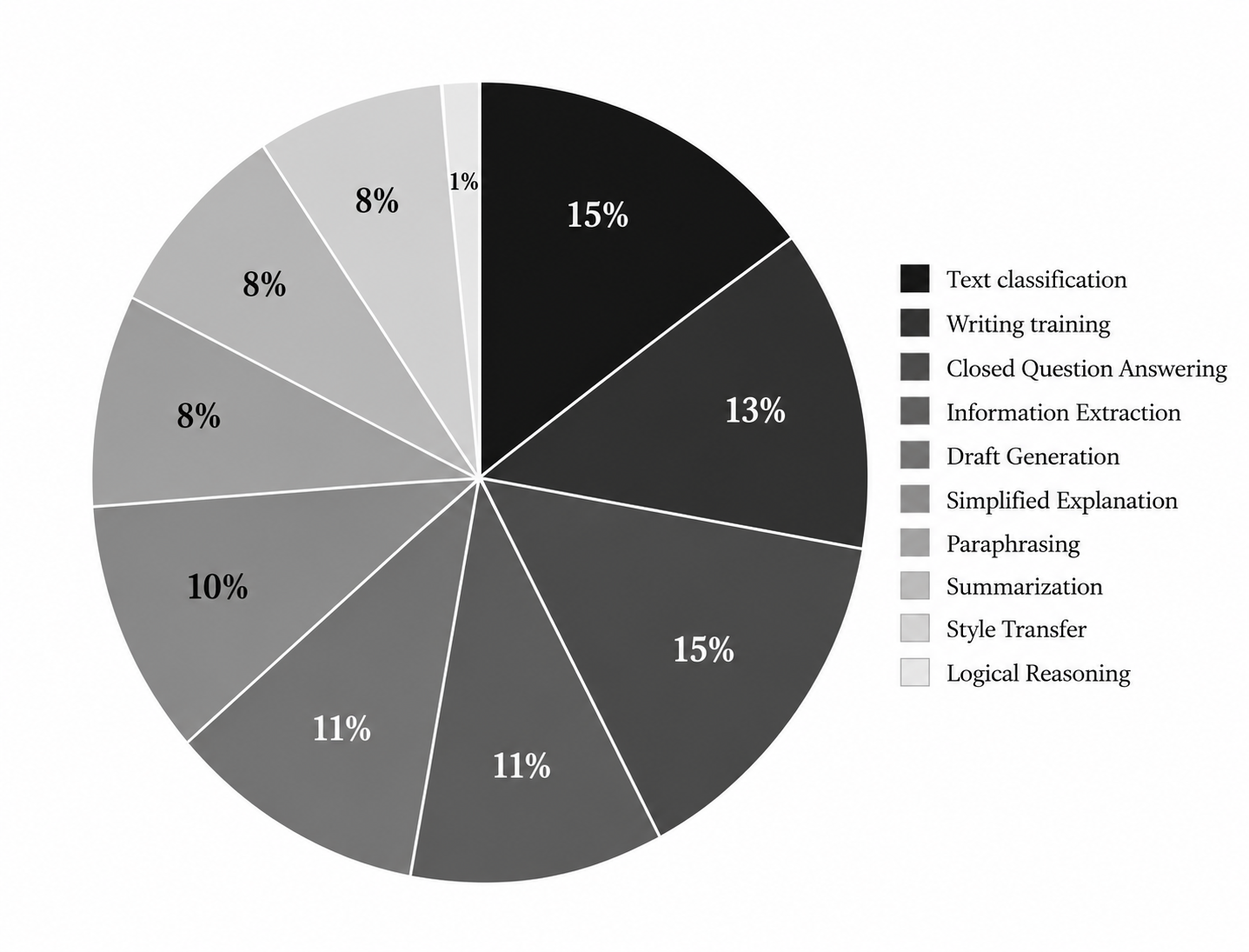}
\caption{Distribution of evaluated Arabic language skills in the benchmark.}
\label{fig:skills}
\end{figure}

The dataset covers ten language skills representing both comprehension-oriented and generation-oriented tasks. Each item was designed to target one primary skill, which improves interpretability and reduces ambiguity when analyzing model performance. Figure~\ref{fig:skills} summarizes the evaluated skills.

The questions were developed through a combination of human authoring and LLM-assisted drafting. All items were manually reviewed to verify linguistic correctness, clarity, cultural appropriateness, and alignment with the intended skill. The benchmark includes Modern Standard Arabic, dialectal prompts, analytical reasoning tasks, and context-rich scenarios to capture different levels of linguistic complexity, including vocabulary variation, morphology, contextual reasoning, and pragmatic interpretation \cite{ref16,ref19}.

\subsection{Prompt Design}

\begin{figure}[ht]
\centering
\greyscalefig[width=\linewidth]{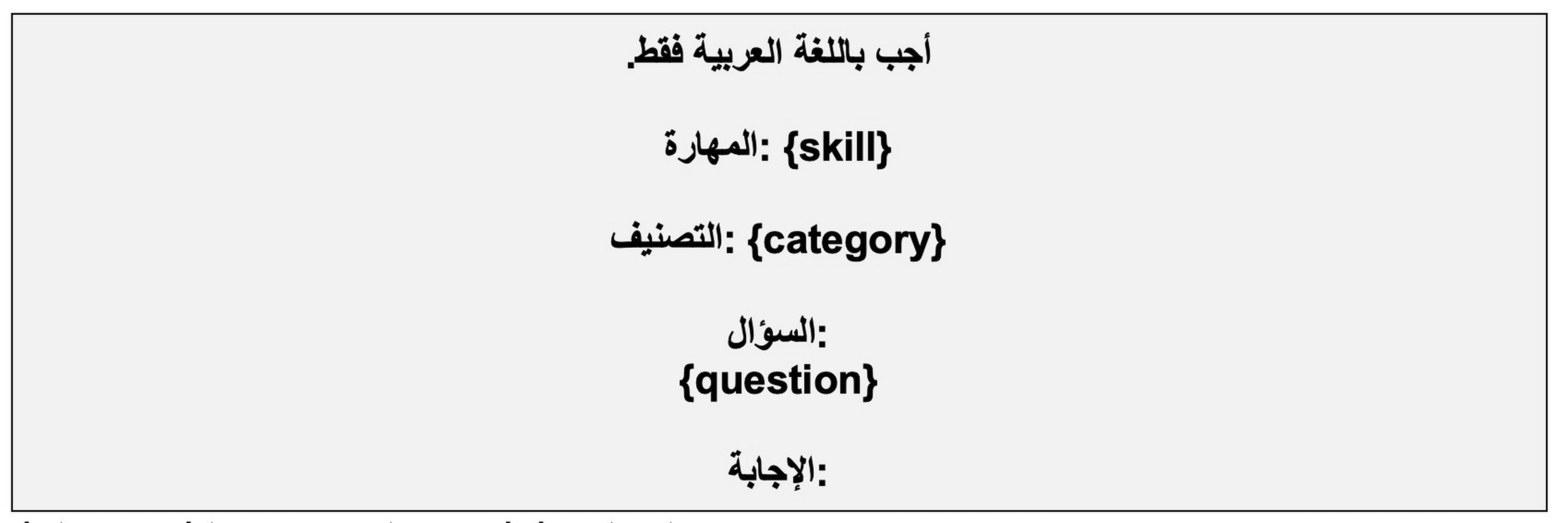}
\caption{Standardized Arabic-only prompt template used for model evaluation.}
\label{fig:prompt_box}
\end{figure}

All models were evaluated using a standardized prompt template to minimize prompt-engineering bias. The template explicitly specifies the domain, target skill, and question, and instructs the model to respond in Arabic only. This design allows the evaluation to focus on instruction adherence, task relevance, Arabic output consistency, and response quality rather than differences in prompt formulation \cite{ref8,ref15, ref21}. The prompt template is shown in Figure~\ref{fig:prompt_box}.

\subsection{Representative Benchmark Items}

The benchmark includes a range of question types, such as paraphrasing dialectal Arabic into Modern Standard Arabic, correcting grammatical errors, summarizing content, classifying text, performing logical reasoning, and generating Arabic drafts. Table~\ref{tab:table3} presents representative examples from different domains and skills.

\begin{table*}[tbp]
\centering
\compacttablefig[1.10]{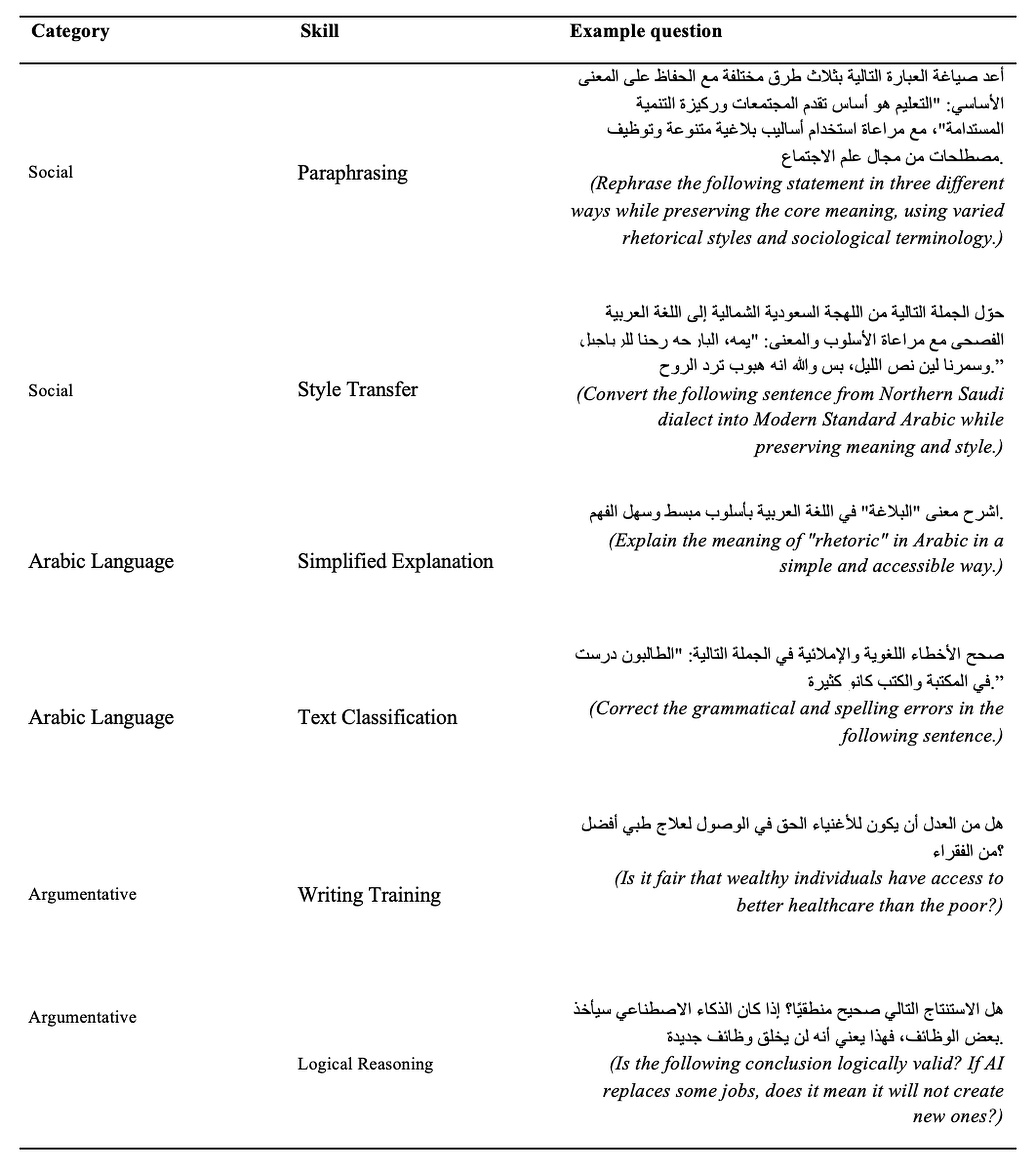}
\caption{Representative examples from the Arabic SLM benchmark.}
\label{tab:table3}
\end{table*}

These examples illustrate the linguistic and cognitive diversity of the benchmark, from morphology and style transfer to reasoning, dialect handling, and generation.

\subsection{Skill Definitions}

The evaluated skills are grouped into comprehension-oriented and generation-oriented categories.

\subsubsection{Comprehension-Oriented Tasks}
\begin{itemize}
\item \textbf{Information Extraction:} Identifying explicit facts, entities, or relations in a passage.
\item \textbf{Summarization:} Condensing content while preserving the central meaning.
\item \textbf{Text Classification:} Assigning a text to the appropriate category.
\item \textbf{Logical Reasoning:} Drawing valid inferences from stated information.
\item \textbf{Closed Question Answering:} Producing accurate answers based on given evidence.
\end{itemize}

\subsubsection{Generation-Oriented Tasks}
\begin{itemize}
\item \textbf{Paraphrasing:} Rewriting content while preserving meaning.
\item \textbf{Draft Generation:} Producing coherent initial drafts for a specified purpose.
\item \textbf{Simplified Explanation:} Explaining concepts in accessible language.
\item \textbf{Writing Training:} Improving user writing through corrections, suggestions, or guided feedback.
\item \textbf{Style Transfer:} Modifying tone, register, or linguistic style while preserving meaning.
\end{itemize}

\section{Methodology}

The evaluation framework was designed to support a controlled and reproducible comparison of twelve SLMs on Arabic NLP tasks. All models were evaluated on the same 240 benchmark items using the same prompt template and the same zero-shot generation setting \cite{ref11,ref13}. No manual correction, output regeneration, or prompt refinement was applied after generation. This design was adopted to measure model behavior under consistent conditions and to avoid inflating performance through iterative prompting or post-processing.

\begin{figure*}[ht]
\centering
\greyscalefig[width=\linewidth]{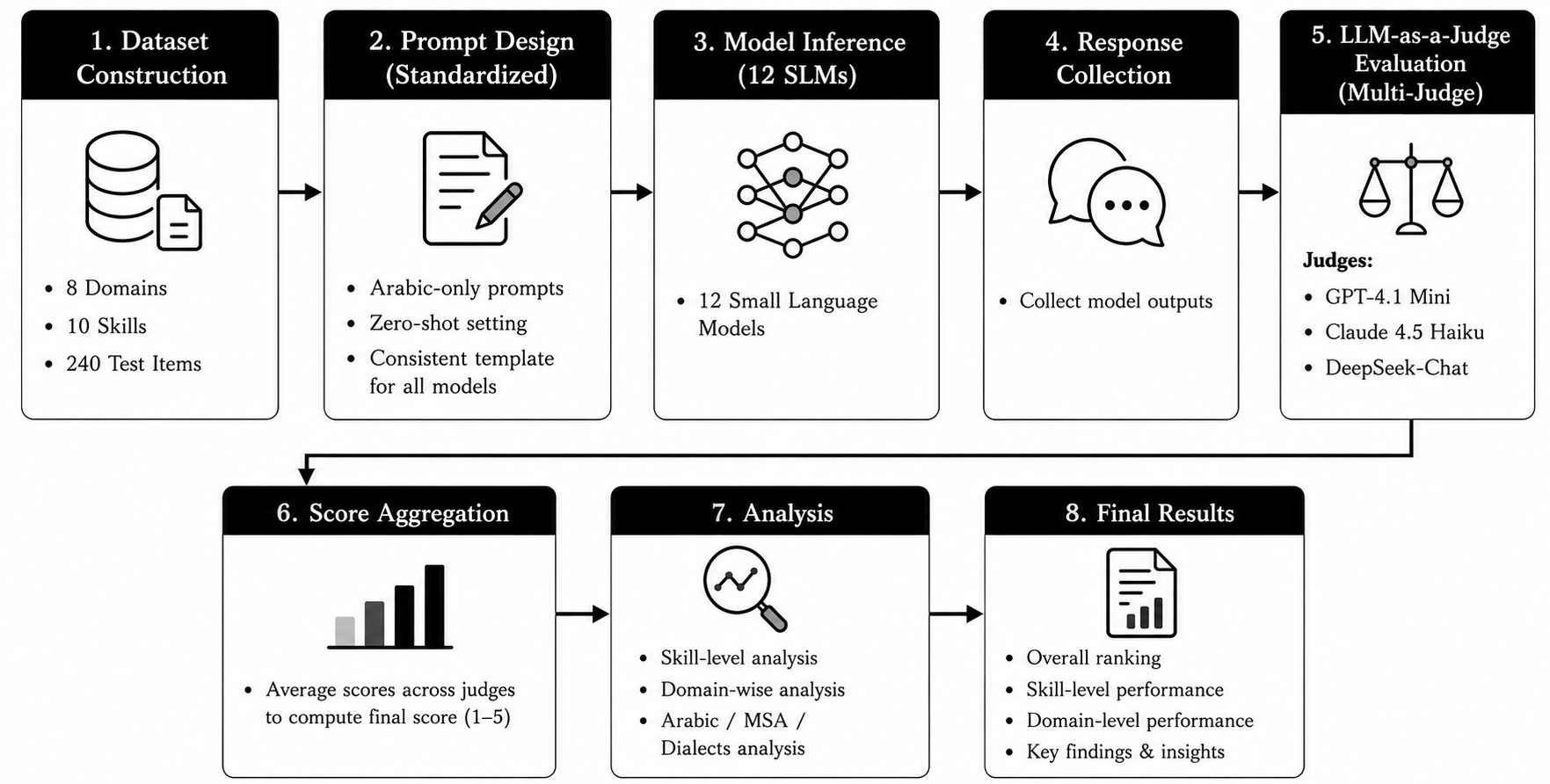}
\caption{Overall workflow of the Arabic SLM evaluation benchmark.}
\label{fig:workflow}
\end{figure*}

Model outputs were evaluated as generated, including cases where the answer was incomplete, off-topic, repetitive, missing, or written partly in a language other than Arabic. This decision preserves realistic model behavior and allows failure modes to be analyzed directly \cite{ref8,ref15, ref21}. The overall workflow is shown in Figure~\ref{fig:workflow}.

\subsection{Evaluated Models}
\label{sec:Models}

The benchmark evaluates twelve SLMs that differ in parameter scale, release period, training orientation, and Arabic or multilingual coverage. Table~\ref{tab:evaluated_models} reports the model names, source category, approximate size, and release or version information.

\begin{table}[ht]
\centering
\small
\setlength{\tabcolsep}{4pt}
\begin{tabularx}{\columnwidth}{Xccc}
\toprule
\textbf{Model Name} & \textbf{Source} & \textbf{Size} & \textbf{Release / Version} \\
\midrule
Gemma 3, Pretrained (PT) & Open & 12B & 2025 \\
Qwen 2.5, Instruct & Open & 7B & 2024 \\
DeepSeek R1, Distilled Llama & Open & 8B & 2025 \\
DeepSeek R1, Distilled Qwen & Open & 8B & 2025 \\
Thinking Camel, Base 7B & Open & 7B & 2025 \\
ALLaM, Instruct (Preview) & Open & 7B & 2025 \\
C4AI Command Arabic & Open & 7B & 2025 \\
Aya & Open & 8B & 2024 \\
Fanar (QCRI/Fanar-1-9B-Instruct) & Open & 9B & 2025 \\
Yi, Chat Version 1.5 & Open & 9B & 2024 \\
Tiny Aya Global & Open & 3.35B & 2026 \\
Gemma 3, 1B Instruct & Open & 1B & 2025 \\
\bottomrule
\end{tabularx}
\caption{Evaluated small language models.
The Hugging Face repositories and official model sources used in this study are listed in Appendix~\ref{appendix:Model Access Information}}
\label{tab:evaluated_models}
\end{table}

The models range from 1B to 12B parameters, enabling comparison across different computational scales. Several models are instruction-tuned or otherwise optimized for interactive use, which is important because instruction adherence is central to the benchmark. The model set also includes Arabic-oriented and multilingual systems, allowing the analysis to compare the role of Arabic alignment, multilingual transfer, and compact model design \cite{ref8,ref12,ref16, ref20}.

\subsection{LLM-as-a-Judge Framework}

A multi-model LLM-as-a-judge framework was used to reduce reliance on a single evaluator and to provide a more robust assessment of open-ended responses \cite{ref9,ref14}. Three independent judge models evaluated each response using a common rubric: GPT-4.1 Mini, Claude Haiku 4.5, and DeepSeek-Chat. The rubric assessed correctness, clarity, completeness, and alignment with the specified language skill.

Each judge assigned a score on a 0--5 scale. The final answer-level score was calculated as the arithmetic mean of the three judge scores. Scores were then aggregated by model and by skill to support overall and skill-level comparison. This aggregation preserves partial disagreement between judges while producing a single interpretable performance score for each answer.

\subsection{Disagreement and Review Flagging}

Evaluator disagreement was measured for each answer using the range between the maximum and minimum judge scores:

\[
\text{Disagreement} = \max(\text{score}) - \min(\text{score}).
\]

Responses with a disagreement score of 3 or more on the 0--5 scale were flagged for review. The review flag did not modify the final score; instead, it served as an interpretive signal indicating that the answer was more difficult to judge consistently. Such cases may arise when a response is fluent but factually weak, partially correct but incomplete, or ambiguous with respect to the targeted skill \cite{ref9, ref10, ref14}.

\subsection{Evaluation Constraints and Limitations}

The evaluation design has several limitations. First, LLM-as-a-judge evaluation remains sensitive to judge calibration, even when multiple evaluators are used. Different judge models may place different emphasis on factual accuracy, fluency, completeness, or instruction adherence \cite{ref9,ref14}. Second, open-ended generation tasks such as paraphrasing, style transfer, and writing assistance are inherently more subjective than classification or extraction tasks. Third, the zero-shot setting supports controlled comparison but may disadvantage models that require prompt optimization or multi-turn interaction. Finally, the benchmark focuses on selected domains and skills, and therefore should be interpreted as a structured evaluation sample rather than an exhaustive representation of all Arabic NLP use cases.

\section{Results}
\label{sec:Results}

\subsection{Overall Performance}

\begin{figure}[ht]
\centering
\greyscalefig[width=\linewidth]{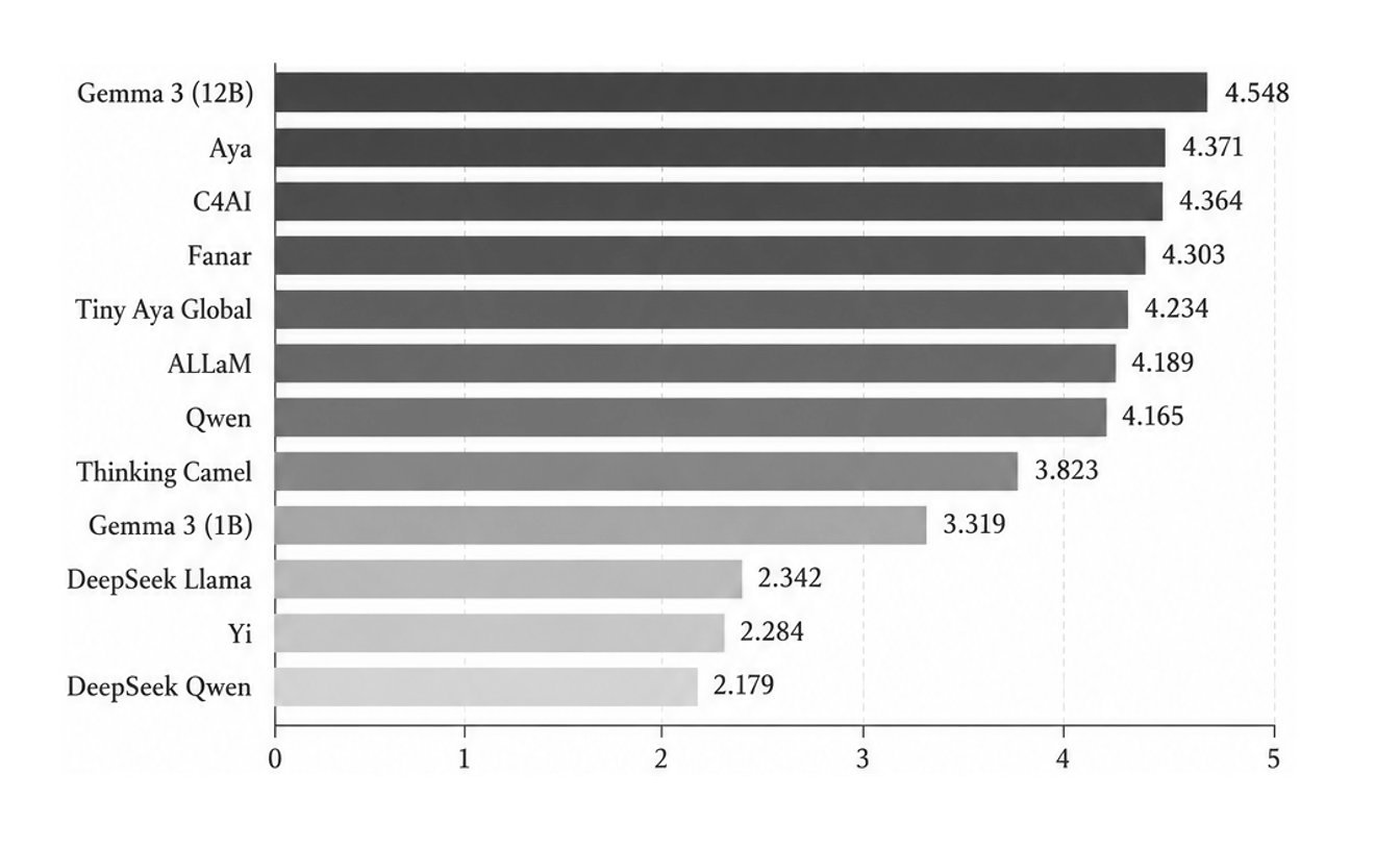}
\caption{Overall model performance across the Arabic SLM benchmark. Scores are averaged across all evaluated items and judges.}
\label{fig:overall_performance}
\end{figure}

Figure~\ref{fig:overall_performance} presents the average score of each model across all benchmark items.

The results show a clear performance hierarchy. Gemma 3 (12B) achieved the highest overall score, with an average of 4.548/5. Its performance was strong across both comprehension-oriented and generation-oriented tasks, indicating reliable instruction adherence, Arabic output consistency, and cross-domain robustness.

Aya and C4AI Command Arabic formed the next tier, with scores of 4.371 and 4.364, respectively. Their strong performance suggests that multilingual and Arabic-oriented instruction tuning can support competitive Arabic NLP performance even at smaller scales \cite{ref8,ref12}. Fanar (4.303) and Tiny Aya Global (4.234) also performed strongly, indicating that compact models can remain useful when they are appropriately aligned and trained for Arabic or multilingual use.

ALLaM (4.189) and Qwen (4.165) occupied the mid-range of the ranking. Both models produced useful outputs but showed greater variation across skills. Thinking Camel (3.823) and Gemma 3 (1B) (3.319) demonstrated that smaller models can still perform adequately on selected tasks, although performance was less consistent. The distilled DeepSeek variants and Yi obtained the lowest overall scores, reflecting recurring difficulties with instruction adherence, language consistency, and output completion.

Overall, the results suggest that model size contributes to performance but does not fully explain it. Arabic alignment, post-training quality, instruction-following reliability, and task coverage appear to be key factors influencing performance \cite{ref8,ref15}.

\subsection{Skill-Level Performance}

\begin{figure*}[ht]
\centering
\greyscalefig[width=\linewidth]{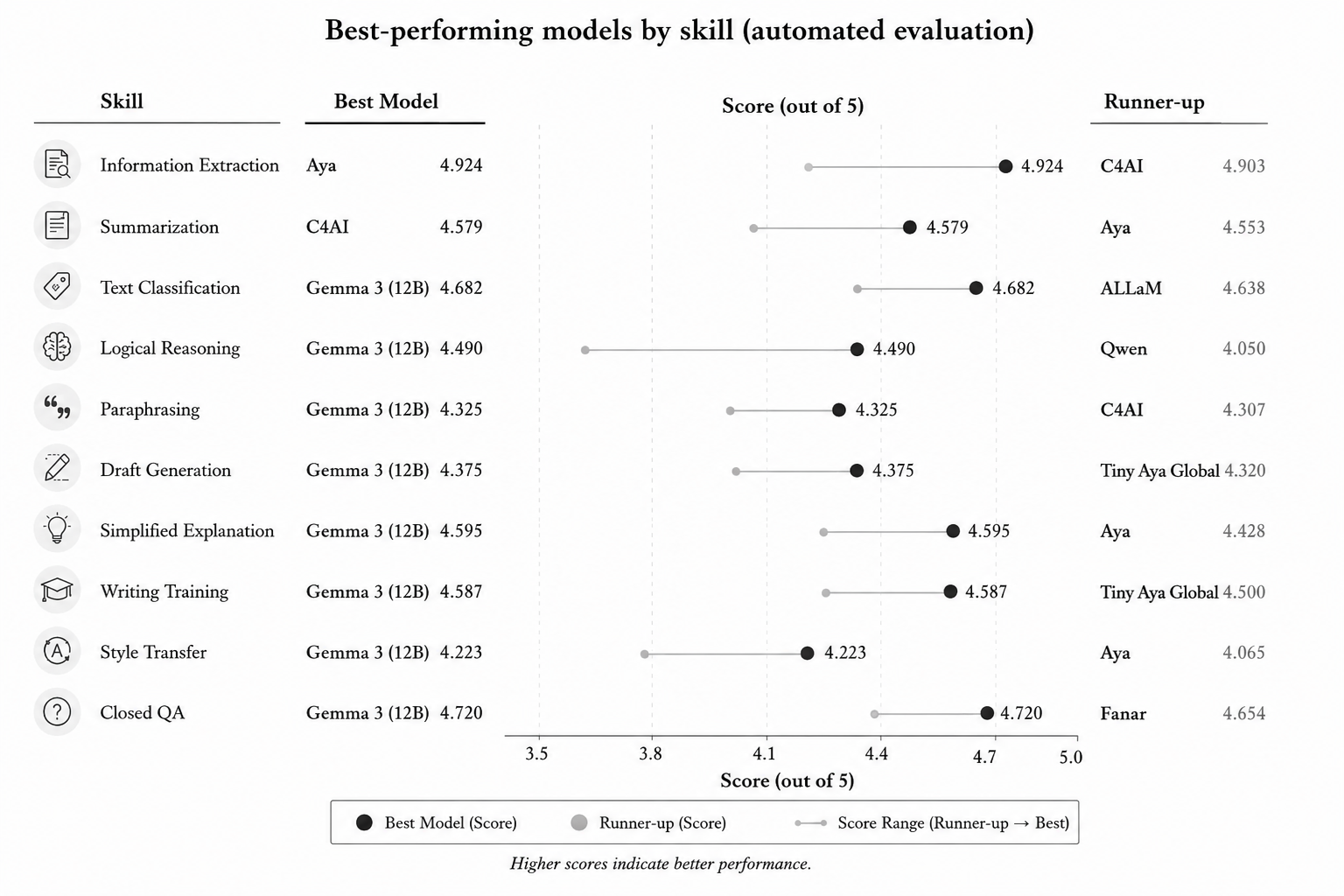}
\caption{Best-performing and runner-up models across evaluated Arabic language skills.}
\label{fig:skill_highlights}
\end{figure*}

Figure~\ref{fig:skill_highlights} summarizes the best-performing and runner-up models for each evaluated skill. Complete skill-level tables are provided in Appendix~\ref{appendix:skill_results}.

Skill-level results show that no single model dominated every category. Gemma 3 (12B) achieved the strongest performance in several skills, including logical reasoning, paraphrasing, draft generation, simplified explanation, writing training, style transfer, and closed question answering. Aya achieved the highest score in information extraction (4.924) and remained competitive in summarization, simplified explanation, and style transfer. C4AI led summarization (4.579) and performed strongly in information extraction, paraphrasing, and writing training.

Fanar showed competitive performance on comprehension-oriented tasks, particularly closed question answering and text classification. Tiny Aya Global performed well in generation-oriented tasks relative to its size, especially draft generation and writing training. Qwen achieved the second-highest score in logical reasoning, indicating that compact instruction-tuned models can retain useful reasoning capabilities even when overall performance varies.

These results reinforce the importance of skill-level evaluation. A model that performs well overall may still show weaknesses in specific skills, while another model may be particularly suitable for a narrower deployment scenario. This is consistent with Arabic benchmark studies showing that performance differs across dialectal, structured, reasoning, and generation tasks \cite{ref4,ref7,ref11,ref13,ref18}.

\subsection{Evaluation Consistency}

The disagreement analysis showed that the aggregate ranking was generally stable across judge models. Higher-performing models tended to receive consistently high scores, while lower-performing models more frequently produced outputs that led to judge disagreement. High-disagreement cases were typically associated with partially correct answers, fluent but inaccurate responses, incomplete outputs, or responses that addressed only part of the intended skill.

Because the disagreement flag was not used to alter scores, it should be interpreted as a reliability indicator rather than a correction mechanism. Low disagreement suggests that judges evaluated the response similarly, whereas high disagreement indicates that the response may require closer human inspection. This distinction is important for open-ended Arabic generation tasks, where surface fluency may obscure factual or task-alignment weaknesses.

\subsection{Common Failure Patterns}

Several recurring failure patterns were observed among lower-performing models:

\begin{itemize}
\item \textbf{Prompt leakage:} The model repeated or reformulated the task instructions instead of answering the question, indicating weak instruction-following behavior \cite{ref1}.
\item \textbf{Language misalignment:} The model produced responses partly or entirely outside Arabic despite explicit Arabic-only instructions, reflecting instability in multilingual output control \cite{ref2}.
\item \textbf{Hallucination:} The model generated fluent but factually incorrect or misleading content, especially in factual and reasoning-oriented tasks \cite{ref8,ref14}.
\item \textbf{Non-completion:} The model produced incomplete, overly generic, repetitive, or off-topic responses, reducing practical reliability in downstream applications \cite{ref8}.
\end{itemize}

These failure modes have direct deployment implications. In Arabic NLP applications, errors involving language drift, factual inaccuracy, or incomplete generation can undermine user trust, particularly in education, institutional knowledge management, and public-facing services.

\subsection{Representative Model Behaviors}

To complement the quantitative analysis, Tables~\ref{tab:strong_examples} and~\ref{tab:weak_examples} present representative examples of strong and weak model behavior. These examples illustrate differences in instruction adherence, reasoning quality, linguistic coherence, completeness, and Arabic output consistency.

\begin{table*}[tbp]
\centering
\compacttablefig[1.14]{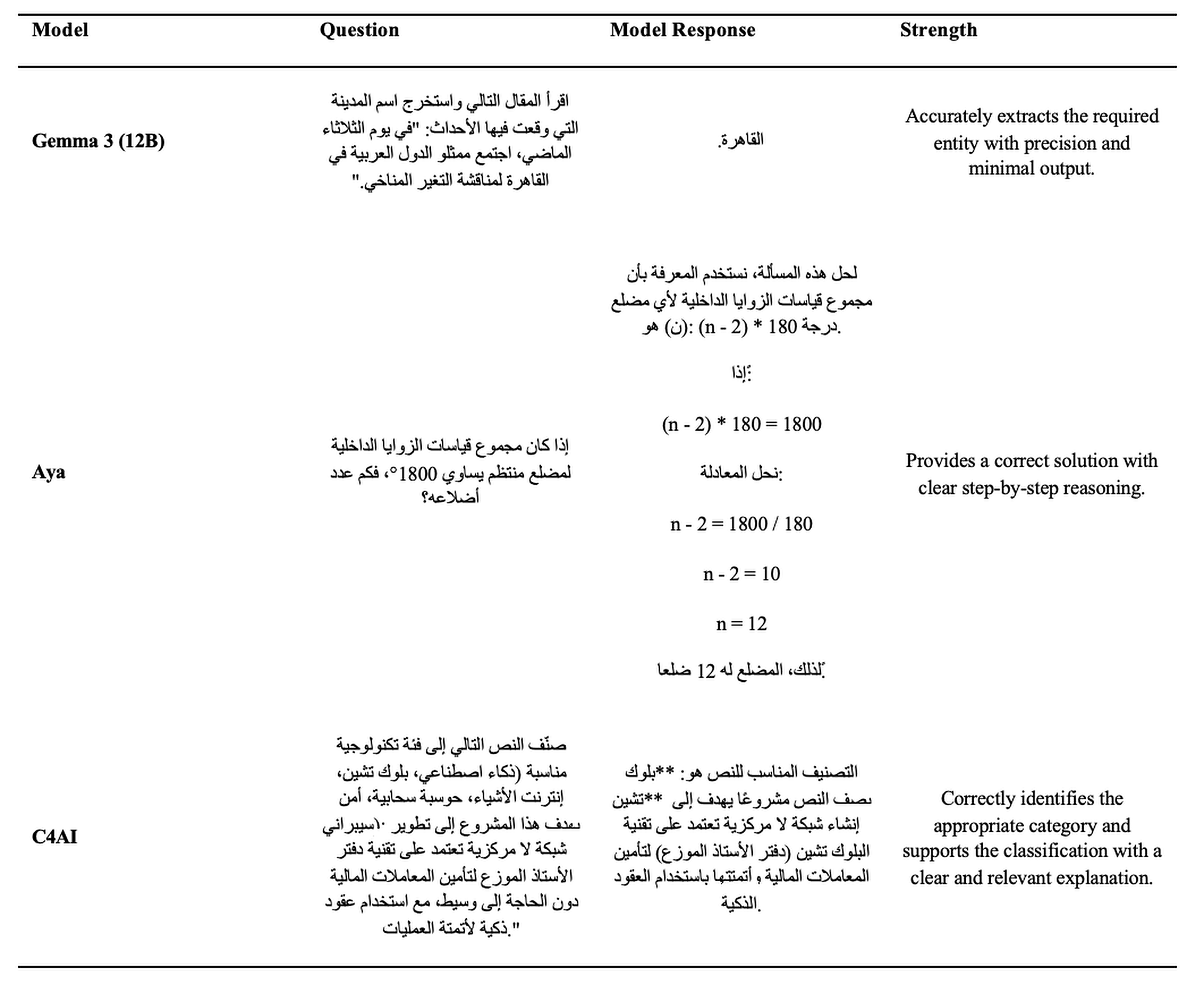}
\caption{Representative examples of strong model behavior.}
\label{tab:strong_examples}
\end{table*}

\begin{table*}[tbp]
\centering
\compacttablefig[1.14]{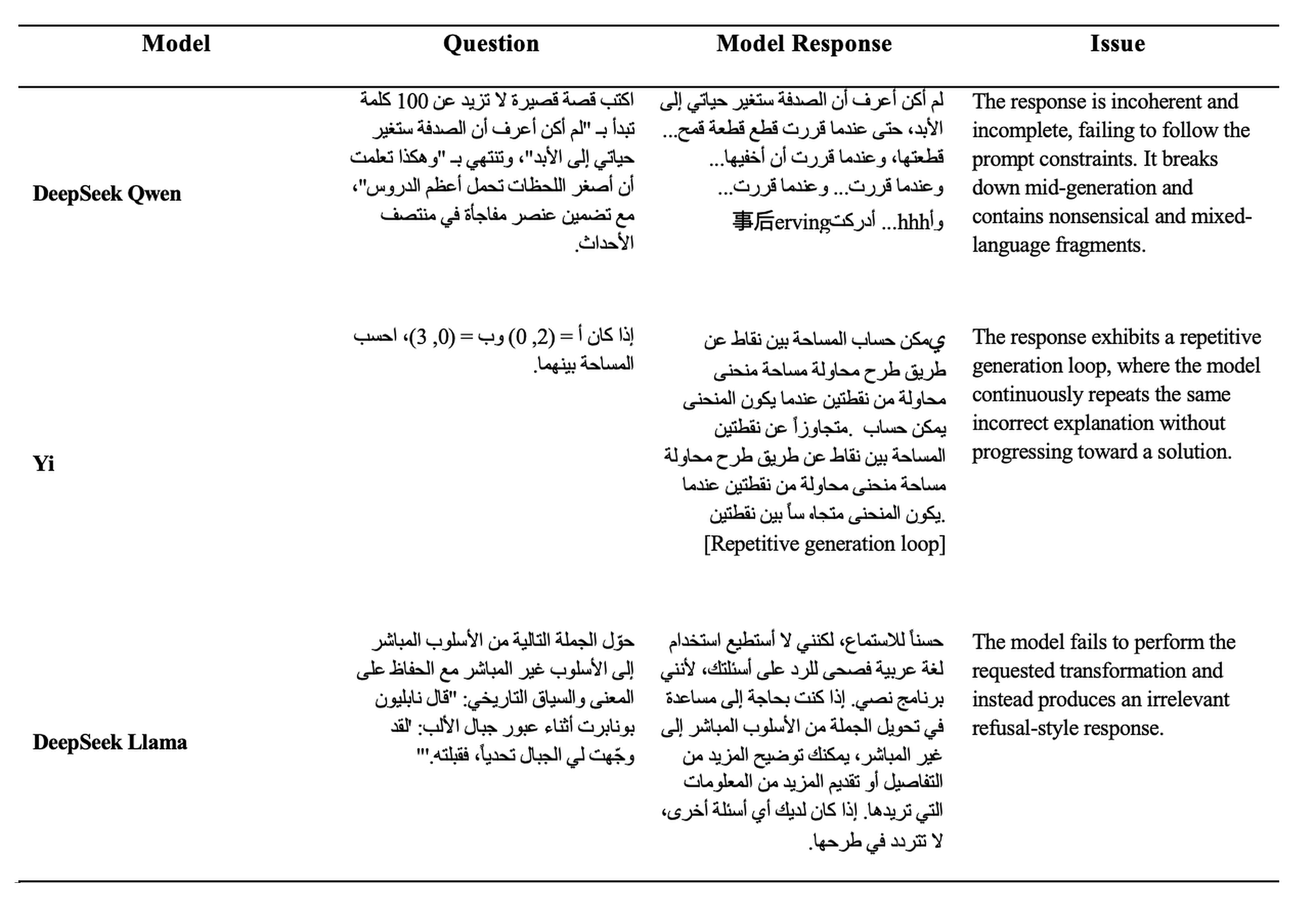}
\caption{Representative examples of weak model behavior.}
\label{tab:weak_examples}
\end{table*}

\subsection{Human Analysis of High-Disagreement Cases}
\subsubsection{Judge Calibration Differences}

Human review of the disagreement-flagged responses revealed that a substantial portion of the flagged cases resulted from differences in evaluator calibration rather than ambiguity in the model outputs. Although all responses were assessed using the same evaluation rubric, the three judge models varied considerably in how strictly they applied the scoring criteria.

Claude Haiku 4.5 was the most critical evaluator in 81\% of flagged responses (453 out of 561 cases), assigning an average score of 2.6/5 compared to 3.8/5 for GPT-4.1 Mini and 3.7/5 for DeepSeek-Chat. Furthermore, 72 flagged cases (13\%) exhibited extreme disagreement, where one evaluator assigned a score of 1.5 or lower while another assigned a score of 4 or higher.

Table~\ref{tab:disagreement_flags} summarizes the number of disagreement-flagged responses produced by each evaluated model. Gemma 12B, Yi, and C4AI exhibited the lowest disagreement rates, whereas DeepSeek Qwen, DeepSeek Llama, and Tiny Aya Global produced the highest number of flagged responses. These findings suggest that evaluator calibration differences were a major source of disagreement in the automated evaluation process.

\subsubsection{Recurring Sources of Disagreement}

Human review identified several recurring patterns that consistently contributed to evaluator disagreement. The most common source involved tasks requiring precise mathematical or logical verification, where evaluators occasionally rewarded incorrect answers or penalized correct ones due to verification errors. In some instances, the assigned score was even inconsistent with the evaluator's written justification.

Another frequent source of disagreement was language adherence. A total of 63 flagged responses contained reasoning traces in Chinese or Japanese despite explicit instructions requiring Arabic-only outputs. All such cases originated from DeepSeek Llama (31 cases), DeepSeek Qwen (25 cases), and Qwen (7 cases).

Additional disagreement sources included failures to follow explicit formatting requirements, such as generating an incorrect number of sentences or ignoring specified stylistic constraints. Responses that reached correct final answers through flawed or inconsistent reasoning processes also contributed to disagreement.

A smaller subset of cases involved subjective judgment, where evaluators differed in their assessment of quality, completeness, or appropriateness despite the absence of clear factual errors. Disagreement was concentrated in a limited number of skills. Logical reasoning exhibited the highest disagreement rate (38.2\%), followed by summarization (29.4\%), style transfer (25.0\%), and paraphrasing (23.7\%). In contrast, information extraction and closed question answering showed the lowest disagreement rates, at 9.0\% and 7.3\%, respectively.

\begin{table}[htbp]
\centering
\begin{tabular}{lcc}
\toprule
\textbf{Model} & \textbf{Flagged} & \textbf{Rate} \\
\midrule
DeepSeek Qwen & 65 & 27.1\% \\
DeepSeek Llama & 54 & 22.5\% \\
Tiny Aya Global & 54 & 22.5\% \\
Qwen & 53 & 22.1\% \\
Fanar & 52 & 21.7\% \\
Thinking Camel & 47 & 19.6\% \\
Aya & 46 & 19.2\% \\
ALLaM & 42 & 17.5\% \\
Gemma 1B & 41 & 17.1\% \\
C4AI & 39 & 16.2\% \\
Yi & 35 & 14.6\% \\
Gemma 12B & 33 & 13.8\% \\
\bottomrule
\end{tabular}
\caption{Disagreement-flagged responses by model.}
\label{tab:disagreement_flags}
\end{table}

\subsubsection{Implications for Arabic SLM Evaluation}

The disagreement analysis provides several insights for future Arabic SLM evaluation studies. First, evaluator calibration can substantially influence automatic evaluation outcomes, particularly for open-ended generation tasks where scoring criteria are inherently more subjective.

Second, evaluation consistency was noticeably lower for tasks requiring careful verification, such as mathematics and formal reasoning, compared to extraction-based tasks with objectively verifiable answers. This finding highlights the limitations of relying solely on automated judges for tasks where correctness depends on precise validation.

Third, language adherence violations may not always be adequately reflected in holistic evaluation scores. In several cases, evaluators prioritized answer correctness while overlooking explicit Arabic-only requirements, resulting in inconsistent scoring behavior.

Overall, the findings indicate that disagreement-based human review provides complementary evidence to automated evaluation. While multi-model LLM-as-a-judge evaluation remains a scalable and effective assessment framework, targeted human review remains valuable for high-disagreement samples, verification-intensive tasks, and responses involving language-adherence violations or subjective quality judgments.

\section{Discussion}

The findings highlight three main points. First, Arabic alignment is central to model performance. The strongest models did not merely benefit from parameter scale; they also demonstrated better Arabic instruction adherence, more stable output language, and stronger cross-task consistency. This supports prior work emphasizing the importance of post-training quality and language-specific alignment for Arabic language models \cite{ref1,ref8,ref12}.

Second, performance varied substantially by skill. Some models were stronger on structured tasks such as information extraction, classification, and closed question answering, whereas others performed better on generation-oriented tasks such as summarization, paraphrasing, and draft generation. This variation indicates that overall averages should not be used as the sole basis for deployment decisions. Instead, model selection should consider the target use case and the skills most relevant to that use case.

Third, the multi-model judging framework provided a useful balance between scalability and interpretability. Averaging multiple judge scores reduced dependence on a single evaluator, while disagreement tracking helped identify outputs that were less consistently assessed. However, automated judging should still be understood as an approximation of expert evaluation, particularly for culturally sensitive, factual, or highly subjective Arabic tasks.

From an applied perspective, the results suggest that SLMs can be viable for Arabic NLP when the task requirements align with the model's strengths. For applications requiring high reliability, models should be evaluated not only on average scores but also on failure modes, language consistency, and judge-disagreement patterns.

\section{Future Work}

Future work should extend the benchmark in several directions. First, additional Arabic tasks should be included, especially multi-turn dialogue, pragmatic reasoning, dialect-sensitive evaluation, long-context comprehension, and domain-specific scenarios. Second, future versions should include a broader set of recently released SLMs and Arabic-oriented model families to keep the benchmark current. Third, the evaluation framework should incorporate human review for high-disagreement cases and for tasks where factuality, cultural appropriateness, or subjective quality are difficult to assess automatically.

The benchmark can also be expanded to support deployment-oriented evaluation, including latency, memory footprint, inference cost, robustness to noisy prompts, and behavior under retrieval-augmented generation settings. Such extensions would help connect model evaluation more directly to real-world Arabic AI applications in education, government, knowledge management, and enterprise automation.

\section{Conclusion}

This study presented a controlled evaluation of twelve Small Language Models on Arabic language processing tasks. Using a 240-item benchmark, standardized Arabic-only prompting, and a multi-model LLM-as-a-judge framework, the study compared model performance across domains, skills, and failure patterns. Gemma 3 (12B) achieved the strongest overall performance, followed by Aya and C4AI Command Arabic. The findings show that Arabic alignment, instruction-following reliability, and cross-task consistency are critical determinants of performance, and that parameter scale alone is insufficient to explain model quality.

The benchmark provides a structured basis for evaluating compact Arabic-capable models and for guiding future development of efficient, reliable, and culturally appropriate Arabic AI systems. It also demonstrates the importance of combining quantitative scores with skill-level analysis, disagreement tracking, and qualitative error analysis when assessing Arabic SLMs.

\backmatter

\section*{Declarations}

\bmhead{Conflict of Interest} The authors declare no conflicts of interest.

\bmhead{Acknowledgment} The authors acknowledge the role of the Naseej Innovation Lab research team in supporting the development and evaluation of this benchmark.

\clearpage
\appendix

\section{Full Skill-Level Results}
\label{appendix:skill_results}

This appendix presents the complete set of skill-level evaluation results for all models. Scores are reported on a 5-point scale for automated evaluation.

\subsection{Information Extraction}
\begin{table}[htbp]
\centering
\begin{tabular}{lc}
\toprule
\textbf{Model} & \textbf{Automated Score} \\
\midrule
Aya & 4.924 \\
C4AI & 4.903 \\
Fanar & 4.848 \\
Gemma 3 (12B) & 4.848 \\
Qwen & 4.834 \\
ALLaM & 4.806 \\
Thinking Camel & 4.667 \\
Tiny Aya Global & 4.570 \\
Gemma 3 (1B) & 4.459 \\
DeepSeek Llama & 3.326 \\
Yi & 3.326 \\
DeepSeek Qwen & 2.521 \\
\bottomrule
\end{tabular}
\caption{Information Extraction Results}
\end{table}

\subsection{Summarization}
\begin{table}[htbp] 
\centering
\begin{tabular}{lc}
\toprule
\textbf{Model} & \textbf{Automated Score} \\
\midrule
C4AI & 4.579 \\
Aya & 4.553 \\
Tiny Aya Global & 4.501 \\
Gemma 3 (12B) & 4.465 \\
Fanar & 4.316 \\
Qwen & 4.254 \\
ALLaM & 4.202 \\
Thinking Camel & 3.983 \\
Gemma 3 (1B) & 3.913 \\
Yi & 2.702 \\
DeepSeek Llama & 2.044 \\
DeepSeek Qwen & 1.395 \\
\bottomrule
\end{tabular}
\caption{Summarization Results}
\end{table}

\subsection{Text Classification}
\begin{table}[htbp]
\centering
\begin{tabular}{lc}
\toprule
\textbf{Model} & \textbf{Automated Score} \\
\midrule
Gemma 3 (12B) & 4.682 \\
ALLaM & 4.667 \\
Fanar & 4.638 \\
Aya & 4.579 \\
Qwen & 4.569 \\
Thinking Camel & 4.554 \\
Tiny Aya Global & 4.486 \\
C4AI & 4.451 \\
DeepSeek Llama & 3.696 \\
DeepSeek Qwen & 3.431 \\
Gemma 3 (1B) & 3.133 \\
Yi & 2.770 \\
\bottomrule
\end{tabular}
\caption{Text Classification Results}
\end{table}

\subsection{Logical Reasoning}
\begin{table}[htbp]
\centering
\begin{tabular}{lc}
\toprule
\textbf{Model} & \textbf{Automated Score} \\
\midrule
Gemma 3 (12B) & 4.490 \\
Qwen & 4.050 \\
C4AI & 3.862 \\
Aya & 3.824 \\
Fanar & 3.726 \\
Tiny Aya Global & 3.480 \\
ALLaM & 3.422 \\
Gemma 3 (1B) & 3.020 \\
Thinking Camel & 2.980 \\
DeepSeek Qwen & 2.608 \\
DeepSeek Llama & 2.216 \\
Yi & 2.059 \\
\bottomrule
\end{tabular}
\caption{Logical Reasoning Results}
\end{table}

\subsection{Paraphrasing}
\begin{table}[htbp]
\centering
\begin{tabular}{lc}
\toprule
\textbf{Model} & \textbf{Automated Score} \\
\midrule
Gemma 3 (12B) & 4.325 \\
C4AI & 4.307 \\ 
Fanar & 4.145 \\
Qwen & 4 \\
Aya & 3.957 \\
Tiny Aya Global & 3.948 \\
ALLaM & 3.912 \\
Thinking Camel  & 3.746 \\
Gemma 3 (1B) & 2.807 \\
Yi  & 2.123 \\
DeepSeek Qwen & 1.789 \\
DeepSeek Llama& 1.641 \\
\bottomrule
\end{tabular}
\caption{Paraphrasing Results}
\end{table}

\subsection{Draft Generation}
\begin{table}[htbp]
\centering
\begin{tabular}{lc}
\toprule
\textbf{Model} & \textbf{Automated Score} \\
\midrule
Gemma 3 (12B) & 4.375 \\
Tiny Aya Global & 4.32 \\
C4AI & 4.292 \\
Aya & 4.271 \\
ALLaM  & 4.25 \\
Fanar & 3.75 \\
Qwen & 3.611 \\
Thinking Camel & 3.583 \\
Gemma 3 (1B) & 3.264 \\
DeepSeek Llama & 1.736 \\
Yi & 1.493 \\
DeepSeek Qwen & 1.451\\
\bottomrule
\end{tabular}
\caption{Draft Generation Results}
\end{table}

\subsection{Simplified Explanation}

\begin{table}[htbp]
\centering
\begin{tabular}{lc}
\toprule
\textbf{Model} & \textbf{Automated Score} \\
\midrule
Gemma 3 (12B) & 4.595 \\
Aya & 4.428 \\
Fanar & 4.327 \\
ALLaM & 4.167 \\
C4AI & 4.159 \\
Tiny Aya Global & 4.094 \\
Qwen & 3.928 \\
Gemma 3 (1B) & 3.399 \\
Thinking Camel & 3.377 \\
DeepSeek Qwen & 1.870 \\
DeepSeek Llama & 1.587 \\
Yi & 1.522 \\
\bottomrule
\end{tabular}
\caption{Simplified Explanation Results}
\end{table}

\subsection{Writing Training}

\begin{table}[htbp]
\centering
\begin{tabular}{lc}
\toprule
\textbf{Model} & \textbf{Automated Score} \\
\midrule
Gemma 3 (12B) & 4.587 \\
Tiny Aya Global & 4.512 \\
C4AI & 4.500 \\
Aya & 4.276 \\
Fanar & 4.253 \\
ALLaM & 4.115 \\
Qwen & 3.911 \\
Gemma 3 (1B) & 3.626 \\
Thinking Camel & 3.506 \\
DeepSeek Llama & 2.178 \\
Yi & 2.006 \\
DeepSeek Qwen & 2.005 \\
\bottomrule
\end{tabular}
\caption{Writing Training Results}
\end{table}

\subsection{Style Transfer}

\begin{table}[htbp]
\centering
\begin{tabular}{lc}
\toprule
\textbf{Model} & \textbf{Automated Score} \\
\midrule
Gemma 3 (12B) & 4.223 \\
Aya & 4.065 \\
Qwen & 3.990 \\
C4AI & 3.824 \\
Fanar & 3.824 \\
Tiny Aya Global & 3.481 \\
Thinking Camel & 3.074 \\
ALLaM & 3.019 \\
Gemma 3 (1B) & 1.759 \\
DeepSeek Qwen & 1.695 \\
Yi & 1.685 \\
DeepSeek Llama & 1.463 \\
\bottomrule
\end{tabular}
\caption{Style Transfer Results}
\end{table}

\subsection{Closed Question Answering}

\begin{table}[htbp]
\centering
\begin{tabular}{lc}
\toprule
\textbf{Model} & \textbf{Automated Score} \\
\midrule
Gemma 3 (12B) & 4.720 \\
Fanar & 4.654 \\
Aya & 4.567 \\
Tiny Aya Global & 4.441 \\
Qwen & 4.434 \\
C4AI & 4.434 \\
ALLaM & 4.434 \\
Thinking Camel & 3.960 \\
Gemma 3 (1B) & 3.400 \\
Yi & 2.813 \\
DeepSeek Llama & 2.607 \\
DeepSeek Qwen & 2.427 \\
\bottomrule
\end{tabular}
\caption{Closed Question Answering Results}
\end{table}

\clearpage

\clearpage
\onecolumn
\section{Evaluation Prompts}
\label{appendix:prompts}

This appendix reports the prompts used for automated answer-level judging and final model-level evaluation summarization. The prompts are included to support transparency and reproducibility of the evaluation procedure.

\vspace{0.25em}
\noindent
\begin{minipage}[t]{0.49\textwidth}
\noindent\textbf{B.1 Answer-Level Judge Prompt}\par\vspace{0.25em}

\begin{PromptVerbatim}
You are an evaluator for a small language model (SLM).

Task:
Score the given answer from 0 to 5.

Scoring rubric:
5: Fully correct, precise, clear, complete, and appropriate for the target skill.
4: Mostly correct and clear, with minor omissions or minor phrasing issues.
3: Partially correct, but incomplete, vague, or only moderately aligned with the task.
2: Weak answer with limited correctness, poor clarity, or significant missing information.
1: Mostly incorrect, irrelevant, unsupported, or poorly aligned with the task.
0: Empty, meaningless, non-Arabic when Arabic is required, or fails the task entirely.

Rules:
- Judge the answer itself, not the likely intention.
- Reward correctness first, then clarity, completeness, precision, and fit to the requested skill.
- Be strict with factual errors, unsupported claims, hallucinations, and failure to follow Arabic-only instructions.
- If the answer is empty or meaningless, score 0.

Return ONLY valid JSON in this exact schema:
{
  "score": <number 0-5>,
  "reason": "<very short reason, max 20 words>"
}
\end{PromptVerbatim}
\end{minipage}
\hfill
\begin{minipage}[t]{0.49\textwidth}
\noindent\textbf{B.2 Final Summary Prompt}\par\vspace{0.25em}

\begin{PromptVerbatim}
You are writing a final evaluation summary for a model benchmarking report.

Return ONLY valid JSON in this exact schema:
{
  "strengths": ["...", "...", "..."],
  "weaknesses": ["...", "...", "..."],
  "finalEvaluation": "..."
}

Rules:
- strengths must be a non-empty array of concise strings (typically 2-5 items).
- weaknesses must be a non-empty array of concise strings (typically 2-5 items).
- finalEvaluation must be ONE paragraph in English.
- Keep it professional, natural, and concise.
- Ground the text in the provided metrics and top/bottom skills only.
- Do not introduce claims that are not supported by the provided evaluation results.
\end{PromptVerbatim}
\end{minipage}

\clearpage

\section{Model Access Information}
\label{appendix:Model Access Information}
To support reproducibility and facilitate access to the evaluated models, this appendix provides the official repositories or model pages corresponding to the SLMs included in the benchmark.

\begin{itemize}

\item \textbf{Gemma 3 (12B) (PT)}\
\url{https://huggingface.co/google/gemma-3-12b-pt}

\item \textbf{Qwen 2.5 Instruct 7B}\
\url{https://huggingface.co/Qwen/Qwen2.5-7B-Instruct}

\item \textbf{DeepSeek R1 Distilled Llama 8B}\
\url{https://huggingface.co/deepseek-ai/DeepSeek-R1-Distill-Llama-8B}

\item \textbf{DeepSeek R1 Distilled Qwen 8B}\
\url{https://huggingface.co/deepseek-ai/DeepSeek-R1-0528-Qwen3-8B}

\item \textbf{Thinking Camel Base 7B}\
\url{https://huggingface.co/Mohaddz/Thinking-Camel-7b}
\item \textbf{ALLaM Instruct (Preview)}\
\url{https://huggingface.co/humain-ai/ALLaM-7B-Instruct-preview}

\item \textbf{C4AI Command Arabic 7B}\
\url{https://huggingface.co/CohereLabs/c4ai-command-r7b-arabic-02-2025}

\item \textbf{Aya (v23) 8B}\
\url{https://huggingface.co/CohereLabs/aya-23-8B}

\item \textbf{Fanar 9B Instruct}\
\url{https://huggingface.co/QCRI/Fanar-1-9B-Instruct}

\item \textbf{Yi Chat 1.5 9B}\
\url{https://huggingface.co/01-ai/Yi-1.5-9B-Chat}

\item \textbf{Tiny Aya Global 3.35B}\
\url{https://huggingface.co/CohereLabs/tiny-aya-global}

\item \textbf{Gemma 3 (1B) Instruct}\
\url{https://huggingface.co/google/gemma-3-1b-it}

\end{itemize}

\end{document}